\documentclass{article}
\usepackage{spconf,amsmath,graphicx}
\usepackage{graphicx}
\usepackage{color}
\usepackage{stfloats}
\usepackage{booktabs}
\usepackage{cite}
\usepackage{amsmath,amssymb,amsfonts}
\usepackage{algorithmic}
\usepackage{graphicx}
\usepackage{textcomp}
\usepackage{xcolor}
\usepackage{url}

\usepackage{fancyhdr}

\title{BiRA-Net: Bilinear Attention Net for Diabetic Retinopathy Grading}

\name{Ziyuan Zhao\thanks{* Both authors contribute equally to this work. This work was supported by National Natural Science Foundation of China (No. 61602349, 61773297)}$^{*\dag}$, Kerui Zhang$^{*\dag}$, Xuejie Hao$^{\ddag}$,Jing Tian$^{\dag}$, Matthew Chin Heng Chua$^{\dag}$, Li Chen$^{\sharp,\S}$, Xin Xu$^{\sharp,\S}$}

\address{$^{\dag}$Institute of Systems Science, National University of Singapore, Singapore\\
$^{\ddag}$University of Electronic Science and Technology of China, China\\
$^{\sharp}$School of Computer Science and Technology, Wuhan University of Science and Technology, China\\
$^{\S}$Hubei Province Key Laboratory of Intelligent Information Processing and Real-time Industrial System,\\
Wuhan University of Science and Technology, Wuhan, China
}

\begin{document}
%
\maketitle

\thispagestyle{fancy}
\fancyhead{}
\lhead{}
\lfoot{\footnotesize{Copyright 2019 IEEE. Published in the IEEE 2019 International Conference on Image Processing (ICIP 2019), scheduled for 22-25 September 2019 in Taipei, Taiwan. Personal use of this material is permitted. However, permission to reprint/republish this material for advertising or promotional purposes or for creating new collective works for resale or redistribution to servers or lists, or to reuse any copyrighted component of this work in other works, must be obtained from the IEEE. Contact: Manager, Copyrights and Permissions / IEEE Service Center / 445 Hoes Lane / P.O. Box 1331 / Piscataway, NJ 08855-1331, USA. Telephone: + Intl. 908-562-3966.}}
\cfoot{}
\rfoot{}

\begin{abstract}
\textit{Diabetic retinopathy} (DR) is a common retinal disease that leads to blindness. For diagnosis purposes, DR image grading aims to provide automatic DR grade classification, which is not addressed in conventional research methods of binary DR image classification. Small objects in the eye images, like lesions and microaneurysms, are essential to DR grading in medical imaging, but they could easily be influenced by other objects. To address these challenges, we propose a new deep learning architecture, called \textit{BiRA-Net}, which combines the attention model for feature extraction and bilinear model for fine-grained classification. Furthermore, in considering the distance between different grades of different DR categories, we propose a new loss function, called \textit{grading loss}, which leads to improved training convergence of the proposed approach. Experimental results are provided to demonstrate the superior performance of the proposed approach.
\end{abstract}


\vspace{-3mm}
\section{Introduction}
\label{sec:intro}

\vspace{-2mm}
\textit{Diabetic retinopathy} (DR) is one of the most common retinal diseases and it is a primary cause of blindness in humans \cite{Das2018}. It augments the blood pressure in small vessels and consequently influences the circulatory system of the retina and the light-sensitive tissue of the eye. Singapore is one of the countries with the highest prevalence of diabetes mellitus. In a recent screening exercise for DR in Singapore, $28.2\%$ of patients with diabetes have been found to have DR, indicating a prevalent condition in the nation \cite{website1}.

The major challenge for DR diagnosis is that DR is a silent disease with no early warning sign, which makes it difficult for a timely diagnosis to take place. The traditional solution is inefficient, in which, well-trained clinicians can manually examine and evaluate the diagnostic images from the Digital Fondus Photography. This type of diagnosis can take several days depending on the number of doctors available and patients to be seen. Besides this, the result of such a diagnosis varies from doctor to doctor and its accuracy relies greatly on the expertise of practitioners. Moreover, the expertise and equipment required may be lacking in many high rate DR areas.

The aforementioned challenges have raised the need for the development of an automatic DR detection system. In recent years, many research works have been conducted to detect DR automatically with the focus on feature extraction and two-class prediction \cite{pinz1998mapping,silberman2010case,sopharak2009automatic,wu2006adaptive}. These works are effective to some extent but also have several shortcomings. First, the extracted features from photos are hand-crafted features which are sensitive to many conditions like noise, exposedness and artifacts. Second, feature location and segmentation cannot be well embedded into the whole DR detection framework. Moreover, only diagnosing to determine whether or not DR is present rather than diagnosing the severity of DR could not well address practical problems and neither can it provide helpful information to doctors.


Recently, \textit{convolutional neural networks} (CNN) has demonstrated attractive performance in various computer vision tasks \cite{LITJENS201760,shi2018bayesian,zhao2019semi}. In this paper, we adopt the CNN-based architecture to develop a five-class DR image grading approach. In the proposed architecture, we design an attention mechanism for better feature extraction and a loss function, called \textit{grading loss}, for fast convergence. In addition, the bilinear strategy is used here for better prediction in this fine-grained image task. Compared to other state-of-the-art research works in five-class classification, the proposed approach is able to achieve superior classification accuracy performance. The contributions of this paper are summarized as below:
\vspace{-5pt}
\begin{itemize}
  \item A new deep learning architecture BiRA-Net is proposed to tackle the DR grading challenge. It contains an attention mechanism that is designed for better feature learning. Moreover, a bilinear training strategy is used to help the classification of fine-grained retina images.
  \vspace{-5pt}
  \item A new loss function based on log softmax is proposed to measure the model classification accuracy for the fine-grained DR grading issue, and it is verified in the experiments to effectively improve the training convergence of the proposed approach.
\end{itemize}
\vspace{-5pt}
The rest of this paper is organized as follows. First, a brief review of the related work is provided in Section \ref{sec:related}. Then the proposed BiRA-Net is presented in Section \ref{sec:method} and compared with the state-of-the-art approaches in Section \ref{sec:experiments}. Finally, Section \ref{sec:conclusion} concludes this paper.


\vspace{-3mm}
\section{Related Work}
\label{sec:related}

\vspace{-2mm}
Conventionally, most diabetic retinopathy detection methods have focused on extracting the regions of interest such as macula, blood vessels, exudates \cite{pinz1998mapping, wu2006adaptive}, and these approaches have dominated the field of DR detection for years. In recent years, CNN has been used in DR detection \cite{lim2014transformed, wang2015hierarchical, gulshan2016development, ghosh2017automatic} and achieved satisfying results in the binary classification of DR. 


Compared to the binary classification, the classification on DR severity is more important, since it can provide more information that can better help doctors diagnose and make decisions. However, even for experienced doctors, it is still a challenge to diagnose the severity of DR based on complex factors of different characteristics of eyes.

Ever since the California Healthcare Foundation put forward a challenge with an available dataset in Kaggle \cite{website2}, more and more research has been put into investigating a multi-class prediction of DR \cite{pratt2016convolutional,bravo2017automatic,zhou2018multi}. Bravo~\emph{et al.} \cite{bravo2017automatic} explored the influence of different pre-processing methods and combined them using VGG16-based architecture to achieve good performance in diabetic retinopathy grading. However, most research utilizes CNN like a black box which lacks intuitive explanation. It is notable that a \textit{Zoom-in-Net} is proposed to use attention mechanism to simulate a zoom-in process of a clinician diagnosing DR and achieve state-of-the-art performance in binary classification \cite{wang2017zoom}.

\vspace{-3mm}
\section{Proposed BiRA-Net: Bilinear attention net for DR grading}
\label{sec:method}

\begin{figure}[htb]
    \centering
    \includegraphics[scale=0.3]{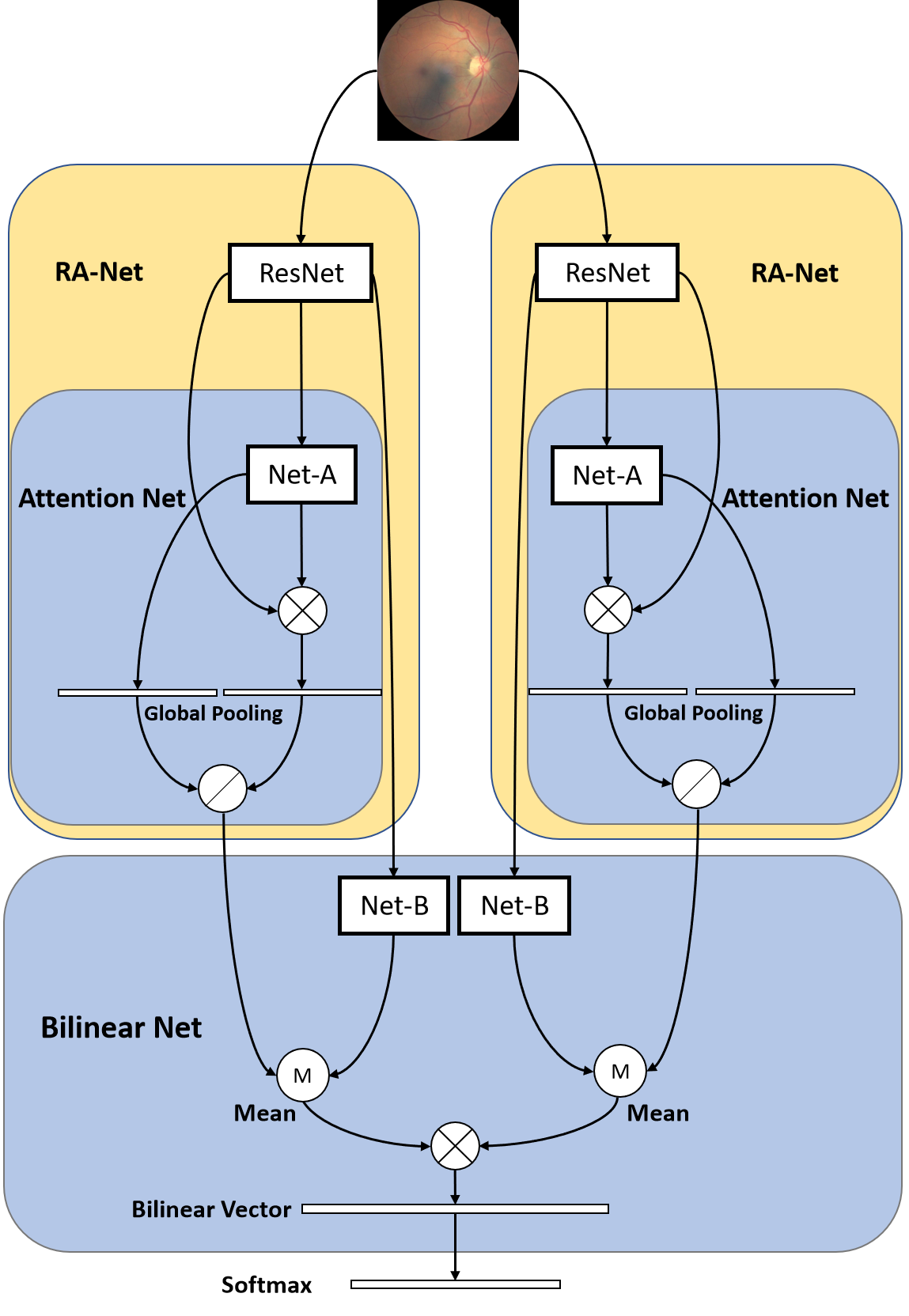}
    \caption{The overview of the proposed BiRA-Net architecture. (A color version of this Figure is available in the electronic copy.)}
    \label{fig:archi}
\end{figure}

\vspace{-2mm}
The proposed BiRA-Net is presented in this Section for DR prediction. The proposed BiRA-Net architecture is shown in Fig.~\ref{fig:archi}, which consists of three key components: (i) ResNet, (ii) Attention Net and (iii) Bilinear Net. First, the processed images are put into the ResNet for feature extraction; then the Attention Net is applied to concentrate on the suspected area. For more fine-grained classification in this task, a bilinear strategy is adopted, where two RA-Net are trained simultaneously to improve the performance of classification. It is for this reason that our architecture is named `` BiRA-Net ".

\vspace{-3mm}
\subsection{Conventional ResNet}

\textit{Residual Neural Network} (ResNet) \cite{resnet} uses shortcut connection to let some input skip the layer indiscriminately, which would avoid adding on new parameters and having too much calculation on the network. Simultaneously avoiding the loss of information and degradation problem, ResNet can saliently increase the training speed and effects. Therefore, the pre-trained ResNet-50 which is 50 layers deep is applied for feature extraction in the proposed network architecture.

\vspace{-3mm}
\subsection{Proposed Attention mechanism}

Medical images always contain much irrelevant information which may disturb decision-making. For our task, microscopic features like lesions and microaneurysms are critical for doctors to classify DR grading. Therefore, the proposed BiRA-Net utilizes the attention mechanism, which mimics the clinician's behavior of focusing on the key features for DR prediction.

\begin{figure*}[htb]
    \centering
    \includegraphics[width=15cm]{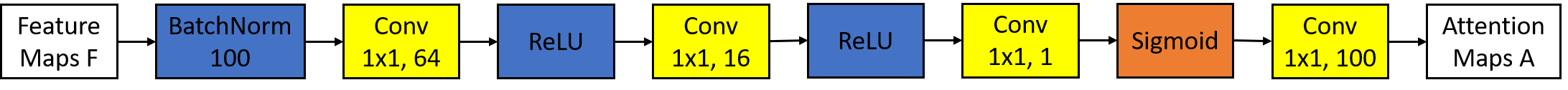}
    \caption{The overview of the attention net (Net-A) used in the proposed BiRA-Net. (A color version of this Figure is available in the electronic copy.)}
    \label{fig:atten}
\end{figure*}

The Attention Net in BiRA-Net firstly takes the feature maps $F \in \mathbb{R}^{100\times 20\times 20}$ from ResNet as input, and then puts them into Net-A which is a CNN network of $3$ convolution layers with 1$\times$1 kernels, which add more nonlinearity and enhance the representation of the network, as shown in Fig.~\ref{fig:atten}, so as to generate attention maps $A \in \mathbb{R}^{100\times 20\times 20}$. Specifically, it produces $20$ attention maps for each disease level by a sigmoid operator. To create the masks $M$ of images, the multiplication between the feature maps $F$ and the attention maps $A$ is applied. And then, we perform \textit{global average pooling} (GAP) on both the masks $M$ and the attention maps $A$ respectively to reduce the parameters and avoid overfitting. Finally, to acquire the weight of images and to filter unrelated information, a division is used. To summarize, the final output for the Attention Net is calculated as
\begin{equation}
\text{Output} = \text{GAP}(A^{l})\oslash \text{GAP}(A^{l}\otimes F^{l})
\end{equation}
where $A^{l}$ and $F^{l}$ are $l$-th attention map and $l$-th feature map, respectively; $\otimes$ and $\oslash$ denote element-wise multiplication and element-wise division respectively.

\vspace{-3mm}
\subsection{Proposed bilinear model}
The proposed BiRA-Net utilizes a bilinear strategy to improve the performance of classification. To speed up the training process and reduce the parameters, two same streams of RA-Net are trained simultaneously, which is inspired by \cite{lin2015bilinear}. More specifically, only one stream needs to be trained, such symmetric bilinear learning strategy has been proved in \cite{kong2017low}.

The Bilinear Net used in the proposed network is illustrated in Fig.~\ref{fig:archi}, which takes the output of Attention Net and the output of the ResNet as inputs. The output of ResNet will be first put into Net-B which is made up of one convolution layer ($100, 20\times20$) and one ReLU activation layer to extract the features and reshape them to be the same as the output of the Attention Net. Then they will be computed by $M$ operator (element-wise mean) as
\begin{equation}
Z^{l}= (X^{l}\oplus Y^{l})\oslash 2
\end{equation}
where $X^{l}$ and $Y^{l}$ are the output of Attention Net and the output of Net-B, $Z^{l}$ is the output of M operator, $\oplus$ denotes element-wise addition. 

Next, we use outer product on the output of $M$ operator term as $Z$ to obtain an image descriptor, and then the resulting bilinear vector $B$ ($B = ZZ^{T}$) is passed through signed square-root step ($Y\leftarrow sign(B)\sqrt{\left | B \right |}$) and $L_{2}$ normalization ($Z \leftarrow Y/\left \| Y \right \|_{2}$) to improve the performance.

\vspace{-3mm}
\subsection{Proposed grading loss}
Conventional loss functions are restricted to reducing the multi-class classification to multiple binary classifications. The distance between different classes is not considered in these conventional loss functions. To reduce the loss-accuracy discrepancy and get an improved convergence, we propose a new loss which adds weights to the softmax function, called `` grading loss ". 

The proposed grading loss function is a weighted softmax with the distance-based weight function and is defined by
\begin{equation}
\mathfrak{L}_{seq}(x,y)=\textup{weight}_{y}\left ( -\textup{log} \left ( \mathfrak{L}_{\textup{softmax}}(x,y) \right )\right)
\end{equation}
where
\begin{equation}\label{formula weight}
\textup{weight}_{y}=\frac{\left | \textup{argmax}(x)-y \right |+1}{M}
\end{equation}
which denotes the softmax function as

\begin{equation}
\mathfrak{L}_{\textup{softmax}}(x,y)= \frac{\textup{exp}(x[y])}{\sum _{j}\textup{exp}(x[j])}
\end{equation}
where
$y\in \left [ 0,C-1 \right ], x=(x_{0},x_{1},x_{2},....,x_{C-1}), M = \sum_{i=0}^{C-1}(\left | y-i \right |+1)$. It defines the gap between class computed by the maximum difference between predicted class $x$ and the real class $y$, and $C$ is the number of class. The weights are normalized by dividing the accumulating of all circumstances as (\ref{formula weight}). 

A toy example is provided to clarify the concept of the proposed grading loss function and it is as follows. For example, in our DR classification task, there are $5$ classes. With the proposed grading loss function, a more substantial price will be paid if it classifies the category with grade $0$ as the category with grade $4$ than that to be the category with the grade $1$, and the weights of these two scenarios are $5/15$ and $1/15$, respectively. This is in contrast to that the conventional loss function imposes a same loss on these two scenarios.

\vspace{-3mm}
\section{Experimental results}
\label{sec:experiments}

\vspace{-2mm}
\subsection{Dataset and implementation}

\begin{figure*}[htb]
    \centering
    \includegraphics[width=14cm]{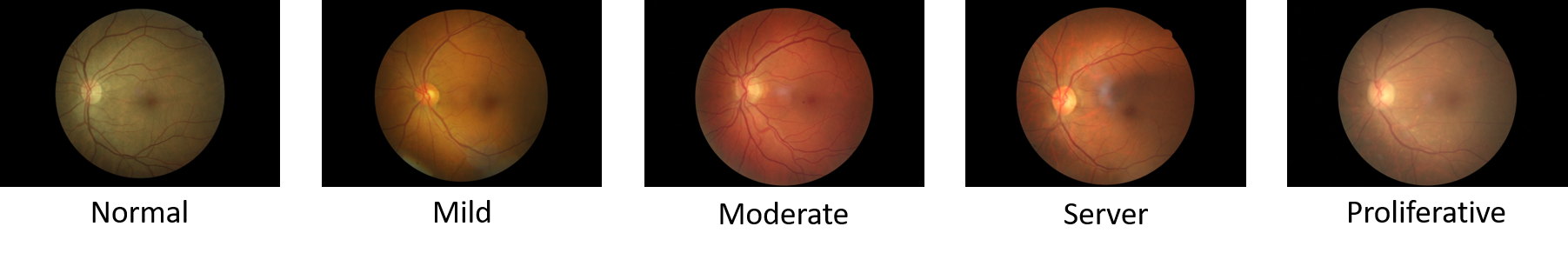}
    \caption{Examples of retinal images \cite{website2} used in the experiment.}
    \label{fig:samples}
\end{figure*}

Experiments are conducted in this paper using a dataset from Kaggle \cite{website2}. The retinal images are provided by EyePACS consisting of $35126$ images. And each image is labeled as $\{0,1,2,3,4\}$, depending on the disease's severity. Examples of each class are shown in Fig.~\ref{fig:samples}. The dataset is highly unbalanced with $25810$ level $0$ images (Normal), $2443$ level $1$ (Mild), $5292$ level $2$ (Moderate), $873$ level $3$ (Server) and $708$ level $4$ (Proliferative). For better generalization and comparison with the latest method of five-class classification, the data distribution from \cite{bravo2017automatic} is adopted, and then we reserve a balanced set of $1560$ images for validation, and the rest are used as training data.



The original images have a black rectangle background. They are cropped to keep the whole retina regions in the square areas. Then the images are resized to $610\times610$ pixels and are standardized by subtracting mean and dividing by standard deviation that is computed over all pixels in all training images. The histogram equalization is used for contrast enhancement. To balance the training data, weighted random sampling is adopted. During training, the images are randomly rotated by $\pm$10 degrees, flipped vertically or horizontally in data augmentation process. 

The proposed model is implemented using Pytorch and trained on a single GTX1080 Ti GPU, using \textit{stochastic gradient descent} (SGD) optimizer with the momentum of $0.9$. The $l_{2}$ regularization is performed on weights with weight decay factor of $5\times e^{-7}$, and the initial learning rate is $0.01$.

\vspace{-2mm}
\subsection{Performance metrics}

\vspace{-1mm}
A confusion matrix is used to count how many images of each class are classified in each class, and the \textit{average of classification accuracy} (ACA) is calculated by the mean of the diagonal of the normalized confusion matrix. Note that an ACA of $0.2$ is the score of a random guess since there are $5$ classes in the experiments.

In addition, the micro-averaged and macro-averaged versions of F1, denoted as Micro F1 and Macro F1 are used to evaluate the results of multi-class classification. F1 is defined as the harmonic mean between precision and recall. The Macro F1 is the mean of the F1-scores of all the classes. In the micro-averaged method, the individual true positives, false positives, and false negatives of different classes are summed up and then applied to get the statistics.

\vspace{-2mm}
\subsection{Baseline methods}
\label{subsec:baseline}

\vspace{-1mm}
We compare our model with the work by Bravo \emph{et al.} \cite{bravo2017automatic}, which has achieved the best ACA using the fusion of VGG-based classifiers with different image preprocessing (circular RGB, grayscale and color centered sets). To explore the effectiveness of various modules in the proposed BiRA-Net, ablation studies are implemented to evaluate the performance of different combinations among different parts as follows.
\vspace{-5pt}
\begin{itemize}
  \item \textbf{Bi-ResNet:} A pre-trained ResNet-50 \cite{resnet} using the proposed bilinear strategy.
  \vspace{-7pt}
  \item \textbf{RA-Net:} Only one single stream of the proposed BiRA-Net is used.
  \vspace{-7pt}
  \item \textbf{BiRA-Net:} The proposed architecture with the proposed grading loss function.
\end{itemize}

\vspace{-5mm}
\subsection{Results}

\vspace{-1mm}


Table \ref{table_results} summarizes the results of all methods on the test dataset. BiRA-Net outperforms all other methods in ACA, Marco F1 and Micro F1. We also implemented BiRA-Net using cross-entropy loss and it has achieved competitive results. The ACA is $0.5424$ which is close to our proposed BiRA-Net. However, using the proposed loss, we observe an improved convergence in speed.

\vspace{-5mm}
\begin{table}[tbh]
\caption{The objective performance evaluation of various approaches, which are described in Section \ref{subsec:baseline}.}
\label{table_results}
\begin{center}
\begin{tabular}{cccc}
\hline
 & ACA  & Marco F1  & Micro F1   \\ \hline
Bravo \emph{et al.} \cite{bravo2017automatic} & 0.5051          & 0.5081          & 0.5052          \\
ResNet-50 \cite{resnet}   & 0.4689          & 0.4753          & 0.4689          \\
Bi-ResNet                                              & 0.4889          & 0.5503          & 0.4897          \\
RA-Net    & 0.4717          & 0.5268          & 0.4724          \\
BiRA-Net                                               & \textbf{0.5431} & \textbf{0.5725} & \textbf{0.5436} \\\hline
\end{tabular}\vspace{-20pt}
\end{center}
\end{table}

\begin{figure}[tbh]
    \centering
    \includegraphics[scale=0.5]{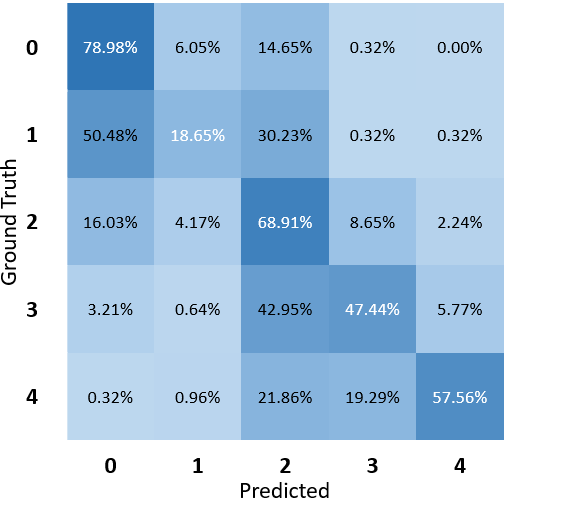}
    \caption{The classification confusion matrix of the proposed BiRA-Net. Horizontal axis indicates the predicted classes and vertical axis indicates the ground truth classes.}
    \label{fig:confusion}
\end{figure}

\vspace{1mm}
Fig.~\ref{fig:confusion} shows the confusion matrix for the proposed BiRA-Net. In the confusion matrix, each class is most likely to be predicted into the right class, except class $1$, which is mostly classified into class $0$. It is clear that class $1$ is the most difficult to differentiate and normal (class $0$) is the easiest to detect.

\vspace{-7mm}
\section{Conclusions}
\label{sec:conclusion}

\vspace{-2mm}
This paper has proposed an attention-driven deep learning architecture for diabetic retinopathy grading, where the bilinear strategy is implemented for fine-grained grading tasks. In addition, the proposed grading loss function helps to attain much improved convergence of the proposed approach. The ablation analyses show that these proposed components effectively improve the classification performance. The proposed BiRA-Net is competitive with the state-of-the-art methods, as verified in the experimental results.

\bibliographystyle{IEEEbib}
\bibliography{refs}

\end{document}